\documentclass{article} 
\usepackage{iclr2026_conference,times}

\usepackage{amsmath,amsfonts,bm}

\def\eqref#1{equation~\ref{#1}}

\def\1{\bm{1}}

\DeclareMathAlphabet{\mathsfit}{\encodingdefault}{\sfdefault}{m}{sl}
\SetMathAlphabet{\mathsfit}{bold}{\encodingdefault}{\sfdefault}{bx}{n}

\usepackage{hyperref}
\usepackage{url}

\usepackage{graphicx}
\usepackage{booktabs}
\usepackage{hyperref}
\usepackage{colortbl}   
\usepackage{subcaption}
\usepackage{pifont}
\usepackage{wrapfig}

\usepackage{tabularx}
\usepackage{booktabs}
\usepackage{multirow}

\usepackage{amsmath}
\usepackage{amsfonts}
\usepackage{amssymb}
\usepackage{makecell}
\usepackage{float}

\usepackage{pifont}

\title{Temporal Aware Pruning for Efficient Diffusion-based Video Generation}

\author{Sheng Li$^{1}$ ~~ Yang Sui$^{4}$ ~~ Junhao Ran$^{3}$ ~~ Bo Yuan$^{3}$ ~~ Yue Dai$^{2}$ ~~ Xulong Tang$^{1}$ \\
$^{1}$University of Pittsburgh \, $^{2}$Illinois Institute of Technology \, $^{3}$Rutgers University \, $^{4}$Rice University \\
}

\iclrfinalcopy 
\begin{document}

\maketitle

\newcommand{\M}{\textsc{xxx}}

\newcommand{\bred}[1]{\textcolor{red}{\sf\bfseries #1}}
\newcommand{\blue}[1]{\textcolor{blue}{#1}}
\newcommand{\yellow}[1]{\textcolor{yellow}{#1}}
\newcommand{\purple}[1]{\textcolor{purple}{#1}}
\newcommand{\brown}[1]{\textcolor{brown}{#1}}
\newcommand{\cross}[1]{\textcolor{red}{\sout{#1}}}

\newcommand{\Yue}[1]{\textcolor{teal}{\textbf{Yue: #1}}}
\newcommand*\circled[1]{\tikz[baseline=(char.base)]{
  \node[shape=circle,draw,fill=black,text=white,font=\bf,inner sep=0.5pt] (char)
  {\small#1};
}}

\newcommand*\circledwhite[1]{\tikz[baseline=(char.base)]{
  \node[shape=circle,draw,fill=white,text=black,font=\bf,inner sep=0.5pt] (char)
  {\small#1};
}}
\begin{abstract}
Video diffusion models have recently enabled high-quality video generation with ViT-based architectures, but remain computationally intensive because generation requires attention computation over long spatiotemporal sequences. 
Token pruning has proven effective for ViTs and VLMs. 
However, most prior pruning methods are attention-based and operate per frame, failing to ensure the vital temporal coherence across frames in video generation tasks.
In practice, naively adopting attention-only pruning causes noticeable degradation due to worsened background consistency, flickering, and reduced image quality.
To address this, we propose TAPE, a training-free Temporal Aware Pruning for Efficient diffusion-based video generation. 
TAPE (i) applies temporal smoothing to align token-importance across adjacent frames and suppress selection jitter; and (ii) performs token reselection in selected layers to align token pruning with layers' diverse semantic focus and avoid error accumulation in specific areas; it also (iii) adopt a timestep-level budget scheduling that prunes aggressively at early noisy steps and relaxes pruning during fidelity-critical refinement. The experimental results show that TAPE delivers significant speedups while preserving high visual fidelity, outperforming prior token reduction approaches.


\end{abstract}    
\section{Introduction}
\label{sec:intro}

Video generation has recently gained tremendous popularity, taking a major role in advancing high-quality video creation~\citep{blattmann2023align, yang2024cogvideox, brooks2024video, kong2024hunyuanvideo}.
The architectures of video diffusion models mainly fall into two categories. Early approaches predominantly adopt UNet-based backbones~\citep{blattmann2023align}, which serve as the foundational framework for video diffusion. To enhance the generation scalability, more recently, DiT-based architectures~\citep{peebles2023scalable} have gained popularity due to strong performance derived from transformer designs.

However, the high computational cost remains a significant challenge due to the inherently long sequence generation in video diffusion models. 
Unlike image-based diffusion models, which handle spatial tokens within a single image only, video diffusion models must simultaneously process spatial and temporal tokens.
Consequently, the overall time complexity increased from $\mathcal{O}((WH)^2)$ (for an image containing $W\times H$ tokens) to $\mathcal{O}((WHT)^2)$ (for a video containing $T$ frames in the size of $W\times H$ tokens). Hence, the computation cost increases sharply with the frame count $T$.
For instance, generating a 5-second video (121 frames at 512$\times$768 resolution with 50 denoising steps) using the HunyuanVideo model~\citep{kong2024hunyuanvideo} takes approximately 14 minutes on an NVIDIA A100 GPU.
This motivates reducing the computational cost without sacrificing visual quality or temporal coherence. 

Recent studies have demonstrated that token pruning can substantially reduce the computational overhead of diverse models, such as Vision Transformers (ViTs) ~\citep{liang2022not, rao2021dynamicvit, kong2022spvit} and Vision-Language Models (VLMs)~\citep{chen2024image, yang2025topv, tao2025dycoke, shao2025holitom, yang2025visionzip, shao2025tokens}. 
By selectively removing less informative visual tokens while retaining those crucial for semantics and structure, these methods achieve significant savings in computation and memory without noticeable accuracy loss.
For example, the recent ViT pruning method can save up to a 24\% FLOP reduction in ViTs with minimal impact on classification performance~\citep{liang2022not}. Similarly, for VLMs, recent work demonstrates up to a 50\% reduction in FLOPs without compromising accuracy~\citep{yang2025topv}. 








Most prior token-pruning methods are \textit{attention-based}: Generally, they remove tokens according to attention-derived importance scores~\citep{kong2022spvit, rao2021dynamicvit, chen2024image, yang2025topv, tao2025dycoke, shao2025holitom}. 
This works well for single-image generation or video understanding tasks, where temporal coherence is either unnecessary or less important.
However, video generation demands strong temporal coherence. Specifically, the frames should change smoothly without flickering, which poses significant demands on frame-to-frame coherence.
As such, it is natural to raise a question: \textit{Do existing token pruning strategies remain effective in diffusion-based video generation?}

To address this inquiry, we systematically evaluated attention-based token pruning for diffusion-based video generation and obtained the following finding:
\textit{Attention-only pruning significantly degrades video quality}.
Using the VBench benchmark~\citep{huang2023vbench}, we observe drops concentrated in background consistency, temporal flickering, and image quality scores.
These results motivate a temporal-coherence-aware token-pruning approach that removes unnecessary computation while preserving cross-frame coherence in video generation.

Therefore, we introduce TAPE, a Temporal Aware Pruning framework for Efficient diffusion-based video generation.
At its core, TAPE applies \textit{temporal smoothing} to align token-importance scores across adjacent frames, suppressing frame-to-frame jitter in token selection.
In addition, TAPE performs \textit{token reselection} in the selected layer to prevent pruning from following a fixed pattern and to allow different regions to be considered throughout the network.
Finally, TAPE adopts a \textit{timestep-level budget scheduling} that relaxes pruning during quality-critical steps, hence saving computation without degrading temporal coherence or visual quality. We summarize our contribution as follows:
\begin{itemize}
    \item We systematically investigate attention-only token pruning for diffusion-based video generation and identify its limits on ensuring temporal coherence.
    \item We propose TAPE, a training-free token pruning framework that couples temporal smoothing of token importance with token reselection and a timestep-level budget scheduling, explicitly preserving cross-frame coherence while reducing computing costs.
    \item Extensive experiments show that TAPE provides significant acceleration with minimal quality degradation. Specifically, TAPE attains a 1.5X speedup while incurring about 1-point decrease in the VBench total score. Compared with existing token-reduction approaches, TAPE consistently achieves superior performance across all pruning rates thanks to its temporally coherent pruning strategy.
\end{itemize}

\section{Background}
\label{sec:background}

\subsection{Diffusion-based Generative Model}
\label{sec:background_diffusion}
Diffusion models have rapidly become the foundation of modern generative modeling, offering exceptional fidelity in visual content synthesis~\citep{ho2022cascaded, epstein2023diffusion}.
At the core of diffusion learning is a stochastic process that gradually transforms random Gaussian noise into structured data through a series of learned denoising steps~\citep{lu2022dpm, ho2020denoising, lipman2023flow}.
Rather than operating directly in the pixel domain of original input data, most diffusion frameworks typically follow the Latent Diffusion Model (LDM) paradigm~\citep{rombach2022high}, where the diffusion process is performed in a compressed latent space encoded by a variational autoencoder.
This latent representation substantially reduces computational cost while preserving fine-grained perceptual detail.

While early diffusion backbones relied on convolutional U-Nets to capture local image structures~\citep{huang2023scalelong, wolleb2022diffusion}, the remarkable success of Transformers in large-scale vision and language modeling~\citep{khan2022transformers, annepaka2025large} has motivated the development of fully attention-based diffusion architectures.
As such, Diffusion Transformers (DiT)~\citep{peebles2023scalable, zhang2025easycontrol} were introduced as a transformer-based backbone for diffusion models, replacing convolutional operations with self-attention over patch tokens.
This design enables long-range dependency modeling and stronger global semantic consistency, and also exhibits superior scalability in large-scale settings~\citep{wang2025lavin, feng2025dit4edit}.

Extending diffusion modeling from images to videos introduces additional challenges stemming from the temporal dimension~\citep{kong2024hunyuanvideo, wan2025wan}.
Unlike static images, videos introduce a temporal dimension where each frame must align coherently with its neighbors in both content and motion.
This temporal dependency significantly complicates the diffusion process: the model must simultaneously reason about appearance, dynamics, and cross-frame continuity.
Moreover, representing videos as sequences of spatiotemporal tokens dramatically increases computational cost, as the number of tokens scales with both spatial resolution and frame count.
As a result, DiT-based video diffusion models face severe challenges in computation cost, motivating the need for an efficient token processing mechanism that can balance generation quality and computation efficiency.

\subsection{Token Reduction for Transformer}
\label{sec:background_token}


Despite the strong modeling capacity, Transformers suffer from the quadratic scaling of self-attention with respect to token sequence length~\citep{dosovitskiy2020image}, which significantly limits the scalability of models such as Diffusion Transformers when handling high-resolution or long-sequence inputs.
This limitation has driven research on token reduction, aiming to reduce computation cost and memory overhead without sacrificing representational fidelity. 
The most popular token pruning strategy is to prune redundant tokens by estimating their importance through attention scores~\citep{liang2022evit, kong2022spvit, xu2022evo, song2022cp, fayyaz2022adaptive, yu2023unified}.
For example, EViT~\citep{liang2022evit} retains the top-$k$ tokens with the highest attentiveness to the [CLS] token and aggregates the rest into a single fused token to preserve global context.
Moreover, SPViT~\citep{kong2022spvit} extends this idea by introducing an attention-based multi-head selector that fuses per-head attention scores, which enables fine-grained token pruning.
Besides the attention-based pruning strategy, STA-ViT~\citep{ding2023prune} introduces a Semantic-aware Temporal Accumulation (STA) score that combines inter-frame similarity and activation strength to prune temporally redundant yet semantically weak tokens.
In parallel, ToMe~\citep{bolya2023tomesd} adopts a merging strategy that fuses tokens with high cosine similarity, effectively reducing computation while preserving global coherence.

Recent work has extended token reduction to Vision–Language Models (VLMs) by leveraging cross-modal attention to identify redundant visual tokens.
FastV~\citep{chen2024image} proposes to discard visual tokens that receive consistently low text-to-vision attention.
However, VisPruner~\citep{zhang2025beyond} observes that cross-attention alone can yield unstable importance estimates due to positional bias; it therefore combines intra-visual attention with token similarity to prune redundant or less informative visual tokens.
Moreover, PuMer~\citep{cao-etal-2023-pumer} integrates text-guided visual pruning based on cross-attention saliency with intra-modal similarity-based token merging to reduce redundancy.
With a focus on video–language understanding task, PruneVid~\citep{huang2025prunevid}, employing a hierarchical pruning strategy that removes temporally static regions, merges spatially similar tokens, and preserves tokens that are most relevant to textual queries, enabling efficient cross-modal reasoning over time.

\section{Does Existing Token Pruning Work Well for Text-to-Video Generation?}
\label{sec:explore}

The most representative token pruning strategy is attention-based token pruning, which evaluates token importance by their attention scores. This method proved to be simple and effective, and has been widely adopted to reduce computation cost while maintaining performance.
However, this method typically focuses on reducing redundant tokens based on attention scores and does not account for the unique challenges posed by text-to-video (T2V) generation tasks, where maintaining temporal coherence across frames is crucial.
For instance, a token that appears unimportant within a single frame might still play a critical role in maintaining continuity across neighboring frames, where temporal coherence is essential.
This raises a natural question: Are existing pruning strategies still suitable for text-to-video generation, where temporal coherence is crucial? To explore this, we extend representative attention-based pruning methods~\citep{liang2022evit} to the video generation setting and analyze the impact on the quality of generated video.

\subsection{Attention-Based Token Pruning}
\label{sec:token_revisit}

Attention-based token pruning has gained widespread popularity due to its simplicity, efficiency, and strong performance across various tasks, as discussed in Section \ref{sec:background}. The core idea of this method is to utilize attention mechanisms, specifically self-attention scores, to assess the relative importance of input tokens. 
Tokens that receive lower attention scores from others are considered less important and more likely to be pruned to reduce the computation costs while maintaining the model's performance.

In Transformer blocks, the self-attention mechanism is employed to process a sequence of tokens. First, each token is projected into three distinct vectors: queries ($Q$), keys ($K$), and values ($V$), which are learned representations. The attention mechanism operates by calculating the similarity between the queries and keys, allowing the model to focus on the most relevant tokens in the sequence. This is done through a scaled dot-product operation, which is defined as:
\begin{equation}
    \text{Attention}(Q, K, V) = \text{Softmax}\left(\frac{QK^T}{\sqrt{d}}\right)V
\end{equation}
Here, $Q$ and $K$ represent the query and key matrices, and $d$ is the dimensionality of the key vectors, used to scale the dot product to maintain the numerical stability of the attention mechanism. The resulting attention scores, derived from the Softmax function applied to the dot product of $Q$ and $K$, determine the weight of each token's contribution to the output. Tokens with lower attention scores contribute less to the output, thus considered less important.

\subsection{Applying Attention-based Token Pruning to T2V Generation}
\label{sec:token_prior}
In this experiment, we adopt an attention-based token pruning strategy similar to EViT~\citep{liang2022evit}, directly leveraging per-token attention scores to identify important tokens. Tokens that receive lower attention scores from others are pruned, while those with higher scores are retained. 
In prior works~\citep{liang2022evit, kong2023peeling}, pruning is applied at selected transformer layers (e.g., the 4th, 7th, 10th blocks of the DeiT model), and once a token is pruned at a given layer, it is discarded and will not appear in subsequent layers. 
However, unlike conventional discriminative tasks such as object recognition, where less important tokens can be safely discarded, generative tasks like video generation require preserving all tokens for reconstruction.
Therefore, importantly, we do not physically discard tokens but skip their computation, maintaining their positional and structural roles so that every token can contribute to the subsequent reconstruction through VAE decoder~\citep{liu2024revisiting}.

We use HunyuanVideo~\citep{kong2024hunyuanvideo} as the baseline model in this study and conduct the experiments on a representative text-to-video generation benchmark, VBench~\citep{huang2023vbench}.
Following prior works~\citep{liang2022evit, kong2023peeling}, token pruning is also applied at selected layers (e.g., the 10th, 30th, and 50th blocks of the HunyuanVideo model, which contains 60 Transformer blocks in total). Once a token is chosen to be skipped, it remains skipped in subsequent layers. 
Also, we apply the same pruning ratio in different temporal frames to avoid frame-wise pruning imbalance.
Table~\ref{tab:token_prelim} presents the quantitative results of applying attention-based token pruning to HunyuanVideo on the VBench benchmark.
The keep rate denotes the ratio of retained tokens after pruning relative to the original token count. 
For example, a keep rate of 80\% indicates that 80\% of tokens are preserved after pruning, and cumulative pruning across multiple blocks further reduces the effective number of active tokens.


\begin{table}[t]
    \footnotesize
    \centering
    \caption{Results of applying the attention-based token pruning approach to the text-to-video generation. The results are obtained on VBench. The HunyuanVideo model is used as bsed model in this experiment.}
    \resizebox{\textwidth}{!}{
    \label{tab:token_prelim}
    \begin{tabular}{m{5.6cm}<{\centering} m{1.6cm}<{\centering} m{2.4cm}<{\centering} m{2.6cm}<{\centering} m{3cm}<{\centering}} 
    \toprule
        Method                                             & Speedup $\uparrow$      & \makecell{Total Score (\%) $\uparrow$}      & Quality Score (\%) $\uparrow$  & Semantic Score (\%) $\uparrow$ \\
        \midrule
        Baseline w/o token Pruning                         & 1.0x          & 83.24                  & 85.09                & 75.82 \\
        Attn-based token prune (keep rate = 80\%)          & 1.4x          & 79.70                  & 81.05                & 74.30 \\
        Attn-based token prune (keep rate = 70\%)          & 1.7x          & 78.25                  & 79.36                & 73.84  \\
        \bottomrule
    \end{tabular}
    }
\end{table}

\begin{table}[t]
    \centering
    \caption{Detailed results of the evaluation dimensions of the quality score. ``kr'' denotes the keep rate. All metrics are higher-the-better, and \textbf{dimensions with significant score drops are highlighted in bold.}}
    \label{tab:prelim_quality}
    \resizebox{\textwidth}{!}{
    \begin{tabular}{l m{2cm}<{\centering} m{2.2cm}<{\centering} m{2cm}<{\centering} m{2cm}<{\centering} m{1.6cm}<{\centering} m{1.6cm}<{\centering} m{1.6cm}<{\centering}}
    \toprule
        Method & Subject Consistency & \textbf{Background Consistency} & \textbf{Temporal Flickering} & Motion Smoothness & Aesthetic Quality & \textbf{Imaging Quality} & Dynamic Degree \\
        \midrule
        Baseline w/o Pruning            & 97.37 & \textbf{97.76} & \textbf{99.44} & 98.99 & 60.36 & \textbf{67.56} & 70.83 \\ 
        Attn-based prune (kr = 80\%)    & 96.18 & \textbf{94.06} & \textbf{96.20} & 97.62 & 59.10 & \textbf{63.48} & 68.61 \\ 
        Attn-based prune (kr = 70\%)    & 96.24 & \textbf{93.26} & \textbf{95.06} & 96.59 & 58.22 & \textbf{61.24} & 68.06 \\ 
        \bottomrule
    \end{tabular}
    }
\end{table}


As shown in the table, although this pruning method improves inference speed, it causes a significant degradation in video generation quality.
With a keep rate of 80\%, the model achieves a 1.4× speedup, but already suffers a noticeable drop in both total and quality scores, decreasing from 83.24 to 79.70 and from 85.09 to 81.05, respectively.
When the keep rate is further reduced to 70\%, the performance degradation becomes more substantial, where the total score drops to 78.25, and the quality score falls by nearly 6 points to 79.36 compared to the baseline.
Although the semantic score also decreases, its change is relatively small compared to the significant loss in visual and temporal quality.
To better understand the cause of this degradation, we further analyze the individual evaluation dimensions that contribute to the quality score.

As shown in Table~\ref{tab:prelim_quality}, this attention-based pruning noticeably affects several key evaluation dimensions of the quality score.
Among them, temporal flickering, background consistency, and imaging quality, experience the most significant drops as the keep rate decreases.
Background consistency declines sharply from 97.76 to 93.26, indicating unstable reconstruction of less-informative background regions across frames.
The temporal flickering score drops from 99.44 to 95.06, revealing visible frame-level incoherence and motion discontinuities.
Imaging quality also decreases from 67.56 to 61.24, suggesting degraded visual clarity in generated frames.

\begin{figure}[t]
    \centering
    \includegraphics[width=0.7\textwidth]{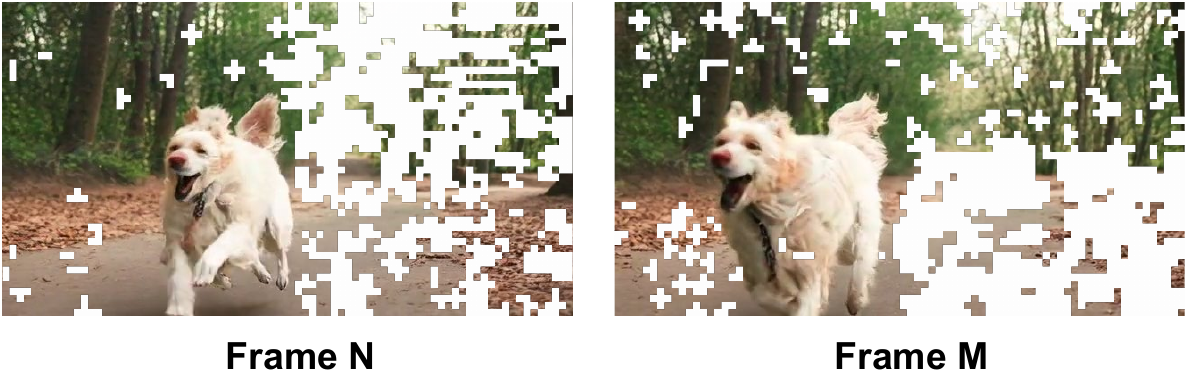}
    \caption{An example to show pruning areas in two frames of a video. The token reduction rate is 30\% in this example.}
    \label{fig:visual_incoherent_pruning}
\end{figure}

The observed degradation across multiple quality dimensions can be attributed to how attention-based token pruning (i.e., token skipping) interacts with the spatiotemporal structure of video generation.
The large drops in background consistency and temporal flickering stem from incoherent pruning decisions across consecutive frames. Because background tokens generally receive lower attention scores, they are frequently pruned; however, the specific background regions being pruned might vary from frame to frame. Even when these regions are indeed low-attention areas, the lack of temporal coherence in the pruning masks leads to visible inconsistencies and flicker. 
As shown in Figure~\ref{fig:visual_incoherent_pruning}, although both frames are pruned in less-informative background regions, the pruned areas differ notably. In Frame N, the pruned regions include both the forest and the road, with a larger proportion located in the forest, whereas in Frame M, the pruning primarily occurs along the road.
This frame-wise incoherence exposes a key limitation of applying attention-based pruning to video generation without considering temporal coherence.

Moreover, the cumulative pruning mechanism in the current strategy significantly affects the score of imaging quality by allowing errors to propagate through deeper layers. In this mechanism, once a token is skipped, it no longer participates in subsequent computations, causing its representation to become outdated and insufficiently refined. When these degraded features are eventually passed to the VAE decoder for frame reconstruction, the accumulated errors result in a significant decline in overall visual quality.

We use attention-based pruning as an example in this preliminary study. However, this issue is not unique to attention-based token pruning; it also exists in other token reduction approaches that overlook temporal coherence in video generation, such as the popular token merging method ToMe~\citep{bolya2023tomesd} and the spatial–temporal adaptive transformer STA-ViT~\citep{ding2023prune}. A more detailed comparison of these approaches is provided in the evaluation section (section~\ref{sec:eval_main}).

\section{Design}
\label{sec:design}



\begin{figure}[t]
    \centering
    \includegraphics[width=\linewidth]{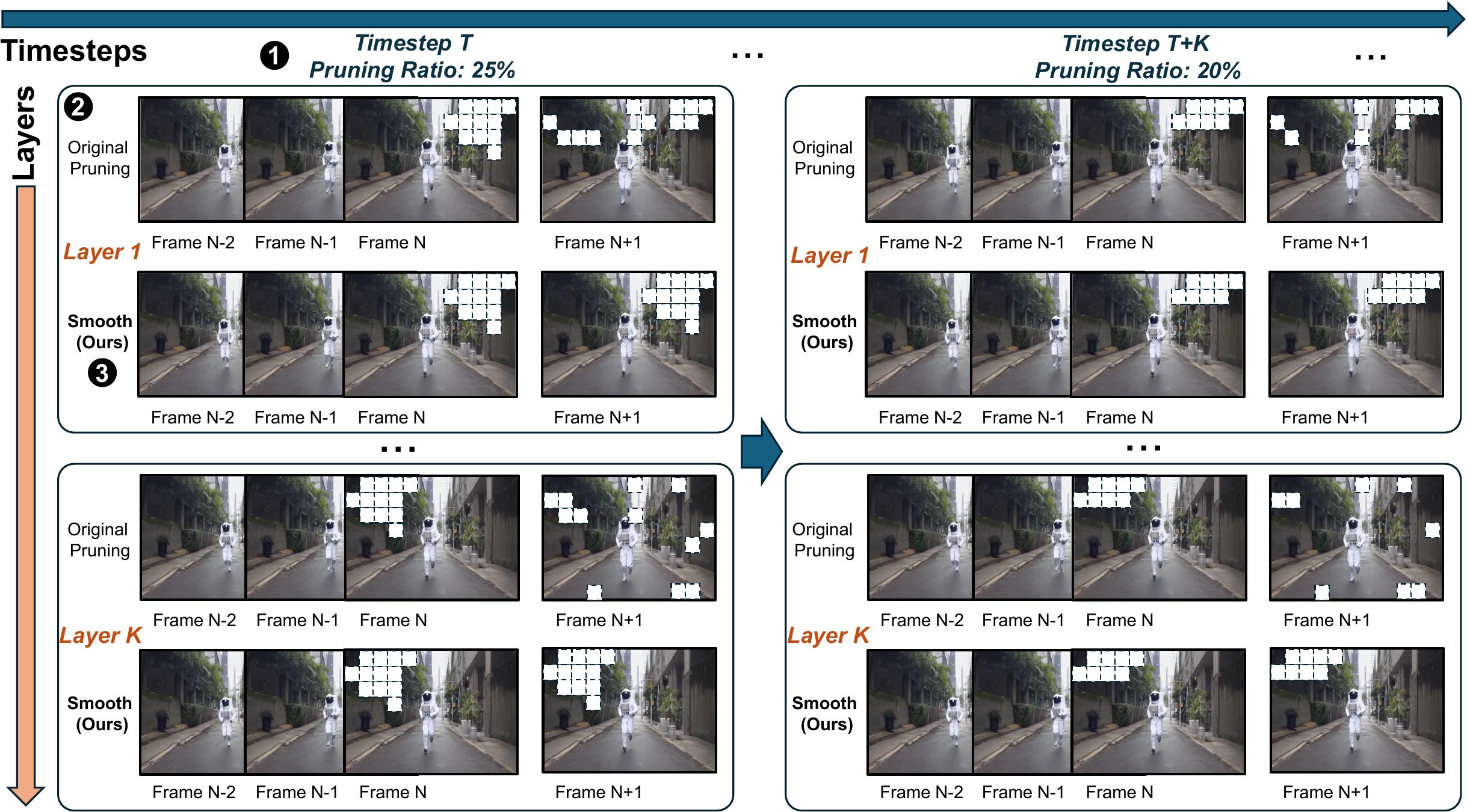}
    \caption{Overview of TAPE. At timestep $T$, \ding{172} timestep-aware scheduling first decides the pruning ratio, which will be reduced at late steps; \ding{173} Token reselection is conducted intermittently, align pruning decisions with diverse semantic focuses in different layers; upon selection, \ding{174} temporal smoothing blends current and aligned previous scores to enforce temporally coherent pruning.}
    \label{fig: design_overview}
\end{figure}

To address the limitations of prior approaches, we propose TAPE, a training-free Temporal Aware Pruning for Efficient diffusion-based video generation. 
At its core, TAPE applies \emph{\textbf{temporal smoothing}} that aligns token-importance scores across neighboring frames to mitigate temporal incoherence in token selection. 
Additionally, TAPE performs \emph{\textbf{token reselection}} at selected layers, allowing tokens with different semantic roles to be revisited and avoiding overly focused pruning.
Moreover, TAPE uses a \emph{\textbf{timestep-level budget scheduling}} that relaxes pruning (i.e., reduces pruning ratio) during fidelity-critical steps, reducing compute without harming temporal coherence or visual quality.

The overview of TAPE is illustrated in Figure~\ref{fig: design_overview}. Given a timestep $T$, TAPE first gets the pruning ratio with the \emph{timestep-level budget scheduling}, which gradually reduces the pruning strengths to let more tokens contribute to the computation in late timesteps. 
Next, \emph{token reselection} revisits token importance at selected layers, ensuring pruning decisions reflect the model’s progressively refined representations. 
The token selection is driven by \emph{temporal smoothing}, which computes smoothed importance by blending each token’s current score with its aligned score from the previous frame, ensuring more temporally coherent pruning than attention-score-based selection in prior works.
Note that, since generative tasks require all tokens for reconstruction, we skip computation for less important tokens rather than physically discard them, ensuring all tokens contribute to the subsequent VAE-based reconstruction.

\subsection{Temporal Smoothing}
\label{sec:design_smoothness}









\textbf{Motivations.}  
While prior attention-based pruning methods ignore temporal coherence, accurately and efficiently modeling it during video generation is non-trivial. 
In particular, avoiding incoherent pruning requires balancing semantic change and spatial/temporal stability. 
On the one hand, the pruning policy must track evolving semantics so that important regions are not mistakenly pruned. 
On the other hand, the pruned region should remain coherent across adjacent frames so that the affected area does not drift unpredictably, which would otherwise induce flicker and disrupt motion coherence.
While Video-analytical solutions, such as optical flow~\citep{cheng2017segflow,tu2022optical} or dense correspondence estimation~\citep{truong2021learning,zheng2022msa}, or learned motion prediction~\citep{bao2019memc, kim2020dynamic}, appear promising for this problem, they fall short for two reasons. 
First, they require accurate physical inputs to recover motion. In contrast, in diffusion-based generation, the input information at most time steps is noisy and often nonphysical, making such signals unreliable or unobservable. 
Second, these methods introduce substantial overheads, such as additional models, training, or per-step motion estimation. These factors can outweigh the computational savings from pruning.
This motivates a lightweight, training-free mechanism that stabilizes token importance scores over time without relying on external motion supervision.

\textbf{Temporal Smoothing Importance Score}. Our design stems from an intuitive observation: \emph{Consecutive frames in videos are strongly correlated.} Specifically, backgrounds, object identities, and coarse geometry change slowly between consecutive frames, and even motion appears as small, continuous changes. 
Consequently, token-importance scores should remain coherent across neighboring frames within the same region.
We therefore propose a \emph{temporal smoothing} scheme that blends each token’s current importance with its previous-frame counterpart, preventing drastic inter-frame swings in scores—and hence abrupt changes in token selection—while preserving genuine motion. 
Specifically, given a token $i$ at frame $n$, we smooth its importance score following Equation \eqref{eq: smooth},
\begin{equation}\label{eq: smooth}
    \tilde{s}^{n}_{i} = \alpha\cdot s_i^n + (1-\alpha)\cdot s_i^{n-1}
\end{equation}
where $s^{\,n}_{i}$ and $s^{\,n-1}_{i}$ are the original importance score (e.g., attention score) at frames $n$ and $n\!-\!1$, respectively, and $\tilde{s}^{n}_{i}$ denotes the smoothed importance score which blends the token's importance score across frames.
The hyperparameter $\alpha\!\in[0,1]$ controls smoothing strength: smaller $\alpha$ gives heavier smoothing (more reliance on the previous frame), and larger $\alpha$ trusts the current frame. According to our practice, we set the $\alpha = 0.5$.
We have conducted a sensitivity study to analyze the choice of $\alpha$ in supplementary material (Section~\ref{sec:evaluation_sensivity}).

By doing so, the smoothed importance score $\tilde{s}^{n}_{i}$ not only weighs the token's importance in the current frame $n$ but also considers its importance in the previous frame $n-1$. 
Consequently, as exemplified in Figure~\ref{fig: design_overview}, tokens selected in the last frame are more likely to be selected again at the current frame, avoiding incoherent pruning across consecutive frames. 


\begin{figure}[!b]
    \centering
    \includegraphics[width=0.8\textwidth]{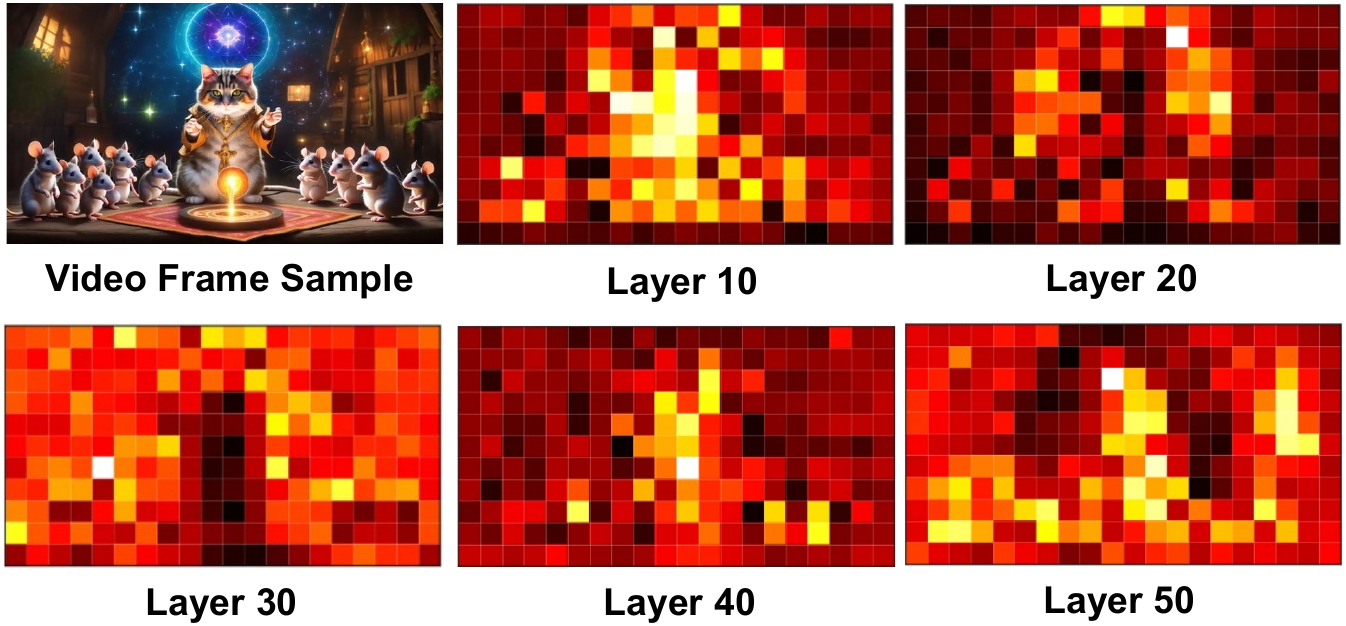}
    \caption{An example of attention distribution across layers. Each block (i.e., token) in the attention map corresponds to the same spatial region of the video frame. And brighter colors indicate tokens that receive higher attention scores from others. The shown attention scores are averaged across all heads.}
    \label{fig:layer-wise-attention}
\end{figure}

\subsection{Token Reselection}
\label{sec:design_layer}





\textbf{Motivation:} It is neither necessary nor desirable to prune the same spatial token across all layers. 
First and foremost, different layers emphasize different semantics. 
As shown in Figure~\ref{fig:layer-wise-attention}, the distribution of attention scores varies across diverse layers. 
As a result, some layers primarily encode stationary structure (layout, coarse geometry), whereas other layers capture dynamic semantics (fine appearance changes and motion cues).
Second, pruning the same set of tokens across diverse layers accumulates errors in that region, leading to noisy outputs such as texture loss and flicker. 
By contrast, token reselection allows pruned tokens to be distributed across wider spatial ranges, thereby avoiding excessive focus on specific areas.


\textbf{Token reselection.} To this end, we propose to intermittently update token importance by recomputing the \emph{smoothed importance scores} (Eq.~\ref{eq: smooth}) for all tokens, re-ranking them, and pruning the least informative ones, which helps prevent pruning from becoming overly biased toward early decisions. 
While more frequent updates improve adaptivity, they also increase computational overhead, as the ``pruned'' tokens need to be brought back into the attention computation more often; in contrast, too few reselections may limit their effectiveness. 
In our practice, we empirically choose to reselect tokens every 10 layers. We provide a sensitivity study on this reselection frequency in the supplementary material (Section \ref{sec:evaluation_sensivity}).



\begin{table}[tb]
	\centering
	\caption{Comparison of different timestep-level pruning schedules under the same attention-based token pruning setting (please refer to Section \ref{sec:token_prior} for details of the attention-based pruning).}
    \label{tab:timestep-level}
    {\footnotesize
	\begin{tabular}{m{2.8cm}<{\centering} m{2.8cm}<{\centering} m{3cm}<{\centering} m{3.2cm}<{\centering}}
	\toprule
	    Method                                         & Total Score (\%) $\uparrow$      & Quality Score (\%) $\uparrow$  & Semantic Score (\%) $\uparrow$ \\
		\midrule
		Constant prune                            & 79.70                 & 81.05          & 74.30 \\
        High$\rightarrow$Low prune                & 81.33                  & 82.85                      & 75.23  \\
        Low$\rightarrow$High prune                & 77.53                  & 78.77                      & 72.58  \\
        \bottomrule
	\end{tabular}
    }
\end{table}

\subsection{Timestep-level Budget Scheduling}
\label{sec:design_timestep}





\textbf{Motivations.} Diffusion proceeds from highly noisy, non-physical latents to increasingly clean and detail-sensitive frames~\citep{lu2022dpm, ho2020denoising, lipman2023flow}. Early in denoising, the signal is coarse and comparatively robust to token removal; late in denoising, fine textures and motion alignment are established, and the model becomes sensitive to drops. To this end, we raise a question: \emph{How should the keep/prune ratio vary along the denoising trajectory to balance efficiency with visual and temporal quality?}

\textbf{Design Space.} To explore the potential effects of changing the pruning budget during video generation, we compare three scheduling policies. (1) \emph{Constant Prune}: We keep prune 20\% tokens at all timesteps.  
(2) \emph{High$\rightarrow$Low Prune}: We allocate 30\% tokens to prune at the beginning, then reduce the value by 5\% every 10 steps, where the total timesteps is 50 in our experiment.   
(3) \emph{Low$\rightarrow$High Prune}: We allocate 10\% tokens to prune at the beginning, then increase the value by 5\% every 10 steps.
Note that all schedules share the same average pruning ratio, and the resulting FLOPs differences are within 1\%, so we regard the computational cost as comparable.
In this experiment, we isolate the effect of the pruning schedule for comparison, by following the attention-based token pruning configuration in Section~\ref{sec:token_prior}, without applying temporal smoothing or token reselection.

As shown in Table~\ref{tab:timestep-level}, we find that \emph{High$\rightarrow$Low} consistently offers the best trade-off: it harvests substantial compute savings during noisy early steps while preserving textures and motion coherence as frames become detail-sensitive. \emph{Constant} yields moderate savings with moderate impact, whereas \emph{Low$\rightarrow$High} tends to degrade late-stage fidelity (texture loss, flicker), indicating that aggressive late pruning harms refinement.

\textbf{Timestep-level Budget Scheduling.} To this end, we adopt a \emph{High$\rightarrow$Low Prune} scheduling scheme. 
Given a target token reduction ratio $\rho$, we set the pruning rate to start from a slightly higher value 
$\rho + \Delta$ and end at a slightly lower value $\rho - \Delta$, where $\Delta$ controls the deviation (e.g., $\Delta=10\%$). The pruning rate is updated at $M$ evenly spaced points along the $T$ denoising steps (e.g., every 10 steps when $T=50$), forming a simple linear decay from $\rho+\Delta$ to $\rho-\Delta$, where the midpoint naturally corresponds to the target ratio~$\rho$. And we set $\Delta=10\%$ by default in our experiments.


\section{Evaluation}
\label{sec:evaluation}

\subsection{Experimental Setup}
\label{sec:eval_setup}


In this section, we evaluate the effectiveness of our training-free temporal-aware pruning framework, TAPE, on text-to-video generation.
Our evaluation is based on the HunyuanVideo model with 13B parameters and 60 DiT blocks.
All experiments are performed directly on the pre-trained model, without any additional training or fine-tuning.
The quantitative comparisons reported in this evaluation section follow the standard evaluation protocol of VBench.
Unless otherwise specified, videos are generated at a 720p resolution.
We have also conducted experiments using another model and metrics, and the results are provided in the supplementary material (Section~\ref{supp:result_wan} and Section \ref{supp:results_metrics}).
All experiments are executed on a single A100 (80GB) GPU.
We compare TAPE to three representative token reduction methods: i) attention-based token pruning method EViT, ii) similarity-based token merging approach ToMe, and iii) STA-ViT, which prunes tokens using a Semantic-aware Temporal Accumulation score that identifies temporally redundant and semantically weak tokens.


\begin{table}[t]
    \setlength\tabcolsep{3pt}
	\centering
	\caption{Comparison of different token reduction approaches on VBench. The full results on all evaluation dimensions of VBench are in the supplementary material (Section~\ref{supp:result_allD}).}
    \label{tab:main_results}
    {\footnotesize
	\begin{tabular}{m{2.2cm}<{\centering} m{2cm}<{\centering} m{1.5cm}<{\centering} m{1.9cm}<{\centering} m{2.4cm}<{\centering} m{2.4cm}<{\centering}}
	           \toprule
               \multirow{2}{*}{\makecell{Token \\ Reduction Rate}}  & Method & Speedup $\uparrow$ & Total Score $\uparrow$ & Quality Score $\uparrow$ & Semantic Score $\uparrow$  \\ 
                        \cline{2-6}
                                                & Baseline                            & 1.0x  & 83.24 & 85.09 & 75.82  \\ 
                                                \hline
        \multirow{4}{*}{ 20\%}                  & EViT           & 1.3x  & 81.82 & 83.62 & 74.64  \\ 
                                                & ToMe         & 1.2x  & 82.18 & 83.92 & 75.22  \\ 
                                                & STA-ViT        & 1.2x  & 80.87 & 82.77 & 73.29  \\ 
                          &  TAPE (Ours)                           & 1.3x  & 83.16 & 84.90 & 76.20  \\ 
                                                \hline
        \multirow{4}{*}{30\%}                     & EViT         & 1.5x  & 81.15 & 82.96 & 73.90  \\ 
                                                    & ToMe     & 1.4x  & 80.92 & 82.74 & 73.66  \\ 
                                                    & STA-ViT    & 1.4x  & 79.30 & 81.30 & 71.28  \\ 
                                                &  TAPE (Ours)     & 1.5x  & 82.01 & 83.64 & 75.50  \\ 
                                                \hline
        \multirow{4}{*}{40\%}                  & EViT            & 1.8x  & 79.38 & 81.53 & 70.77  \\ 
                                            & ToMe             & 1.8x  & 78.26 & 79.88 & 71.78  \\ 
                                              & STA-ViT          & 1.7x  & 77.46 & 79.44 & 69.53  \\
                                                & TAPE (Ours)     & 1.8x  & 81.04 & 83.20 & 72.42  \\ 
        \bottomrule
    \end{tabular}
    }
\end{table}


\subsection{Main Results}
\label{sec:eval_main}

Here we evaluate all the token reduction approaches under different token reduction rates on VBench.
As shown in Table~\ref{tab:main_results}, TAPE delivers significant inference speedups while incurring only minor visual degradation. 
With a more conservative pruning ratio (e.g., 20\%), the VBench score remains nearly unchanged, indicating that our temporal-aware pruning strategy can remove redundant computation without harming generation quality when an appropriate reduction rate is chosen.

When compared with the three representative token-reduction methods, TAPE consistently achieves notably higher scores across all pruning rates.
For example, at a 30\% reduction rate, TAPE attains a total score of 82.01, whereas EViT, ToMe, and STA-ViT fall to 81.15, 80.66, and 79.56, respectively.
This is because all these methods do not explicitly account for temporal coherence, which limits their effectiveness in video generation. 
In contrast, TAPE incorporates temporal smoothing, token reselection, and a timestep-aware pruning schedule, enabling it to maintain cross-frame consistency and preserve visual fidelity while still providing notable acceleration.
We provide a visualization of the pruning areas in some samples in the supplementary material (Section \ref{supp:visual_prune}), to illustrate how our pruning method produces more temporally coherent token selection and avoids frame-wise jitter.

\begin{figure}[tb]
    \centering
    \includegraphics[width=0.85\textwidth]{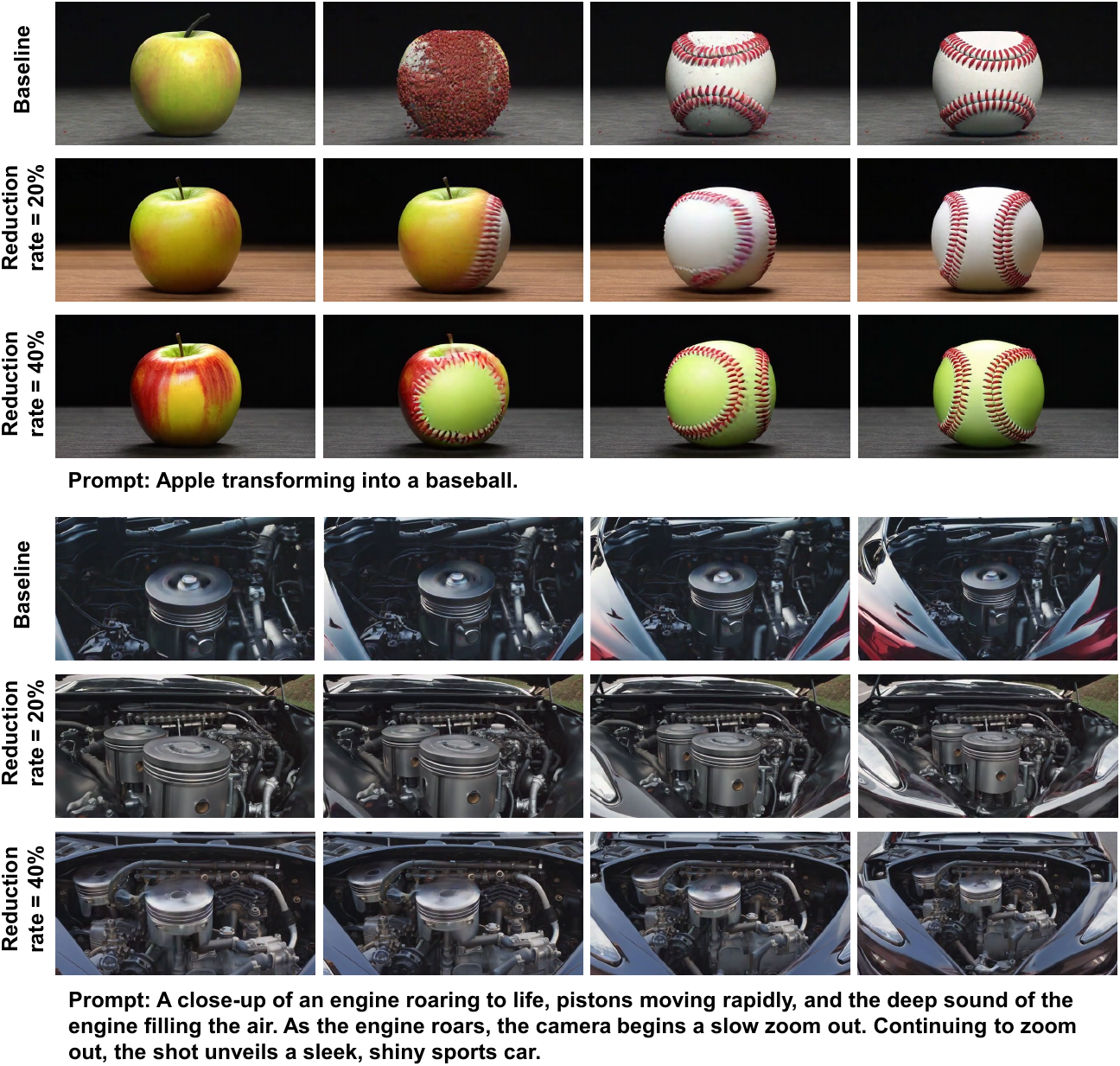}
    \caption{Visualization of the generated video.}
    \label{fig:visual_video_main}
\end{figure}

\subsection{Visualization of Generated Videos}
\label{sec:evaluation_visual_video}

Figure \ref{fig:visual_video_main} presents qualitative examples comparing the baseline model with our TAPE design at 20\% and 40\% token reduction rates. 
Even after pruning a substantial portion of tokens, the generated videos remain highly faithful to the prompts. 
For instance, in the example of ``roaring engine'', the model not only preserves the main subject and the details, but also correctly follows the camera instruction (\emph{zoom out}).
At a 20\% reduction rate, the results are nearly indistinguishable from the baseline. 
Although a 40\% token reduction rate introduces slight softness in some regions, the overall structure, motion, and prompt semantics are still well captured, demonstrating that TAPE maintains strong visual fidelity even under aggressive pruning.
We provide additional visualizations of videos generated with our pruning method TAPE in the supplementary material (Figure \ref{fig:visual_video_supp} in Section \ref{supp:visual_video}), offering a broader view of its visual quality under different scenarios.


\begin{table}[htbp]
\centering
\small
\setlength{\tabcolsep}{3pt}
\caption{Ablation results of \textsc{TAPE} components. We compare \textit{Speedup} (higher is better) and total VBench score. The token reduction rate is 30\%.}
\label{tab:ablation}
    {
    \begin{tabular}{p{4.5cm} m{2.2cm}<{\centering} m{2.8cm}<{\centering}}
    \toprule
    \multirow{1}{*}{Method} & \multirow{1}{*}{Speedup $\uparrow$} & \multirow{1}{*}{Total Score $\uparrow$} \\
    \midrule
    Baseline (no pruning)    & 1.0x & 83.24 \\
    EViT                     & 1.5x & 81.15 \\
    \midrule
    TAPE-S (smoothing only)           & 1.5x & 81.76 \\
    TAPE-R (reselection only)         & 1.5x & 81.52          \\
    TAPE-T (timestep schedule only)   & 1.5x & 81.66           \\
    \rowcolor{white}
    TAPE                & 1.5x & 82.01 \\
    \bottomrule
    \end{tabular}
    }
\end{table}

\subsection{Ablation Study}
\label{sec:evaluation_ablation}
To evaluate the effectiveness of the components separately, we further compare TAPE's performance with each component enabled. Specifically, we compare the performance of TAPE under four versions: (i) \textit{TAPE-S} enables temporal smoothing only; (ii) \textit{TAPE-R} enables token reselection only; (iii) \textit{TAPE-T} enables timestep-aware budget scheduling only; (iv) \textit{TAPE} enables all.
As shown in Table~\ref{tab:ablation}, each component of TAPE provides a clear improvement over the naïve attention-based pruning method EViT.
While smoothing, reselection, and timestep-aware scheduling each offer noticeable gains on their own, none of them matches the performance achieved when all three are combined.
This demonstrates that the three mechanisms are complementary, and together they enable a significantly stronger speed–quality tradeoff.

\section{Conclusion}
In this paper, we presented TAPE, a temporal-aware token pruning mechanism that enables efficient inference for diffusion-based video generation. By enforcing temporal consistency in token selection and adapting pruning across layers and timesteps, TAPE effectively reduces redundant computation while preserving semantic and temporal coherence. Comprehensive evaluations validate its advantage over existing token pruning and merging methods, demonstrating its practicality for accelerating modern video diffusion models without compromising visual fidelity.


\bibliography{main}
\bibliographystyle{iclr2026_conference}

\appendix
\newpage

\section{Sensitivity Study}
\label{sec:evaluation_sensivity}

\subsection{Sensitivity Study on Smoothing Weight}
In this section, we study the sensitivity of our method to the smoothing weight $\alpha$.
We evaluate the total VBench score, speedup, and the three quality-related dimensions where attention-only pruning degrades most as discussed in Section \ref{sec:token_prior} and Table \ref{tab:prelim_quality} (i.e.,\textit{Background Consistency}, \textit{Temporal Flickering}, and \textit{Imaging Quality}) under $\alpha\in\{0.1,0.3,0.5,0.7,0.9\}$.
As shown in {Table~\ref{tab:eval_sens_alpha}, there is a clear trade-off: larger $\alpha$ gives more weight to the current-frame scores and tends to preserve \textit{Imaging Quality} better but yields weaker gains in \textit{Background Consistency} and \textit{Temporal Flickering} (e.g., at $\alpha=0.9$, \textit{Imaging Quality} reaches 66.16, while \textit{Background Consistency} and \textit{Temporal Flickering} drop to 95.32 and 97.73).
Conversely, smaller $\alpha$ emphasizes the previous-frame scores and improves temporal coherence (e.g., at $\alpha=0.1$, \textit{Background Consistency} and \textit{Temporal Flickering} rise to 96.33 and 98.68, but \textit{Imaging Quality} decreases to 64.52).
Overall, $\alpha=0.5$ achieves the best balance---with the highest total score and consistently strong performance across dimensions---so we adopt $\alpha=0.5$ by default.

\subsection{Sensitivity Study on Token Reselection Frequency/Times}
We further study how often to perform token reselection across layers. Specifically, we vary the reselection times over the 60 attention blocks of Hunyuan-Video while fixing the token reduction rate to 30\%, and report the VBench scores and speedups in Table~\ref{tab:eval_sens_freq}. As the reselection becomes more frequent, the model can better adapt token choices to different layers’ semantic focus, which generally improves quality. When we increase the reselection times from 2 to 3 and then to 5, the total score rises from 81.27 to 81.51 and 82.01, with both quality (from 83.07 to 83.26 to 83.64) and semantic scores (from 74.08 to 74.53 to 75.50) improving consistently. 
However, pushing the frequency further to 10 times (every 5 blocks) brings only marginal gains over 5 times (e.g., total score 82.04 vs.\ 82.01, quality 83.68 vs.\ 83.64, semantic 75.48 vs.\ 75.50) while reducing the speedup from 1.5$\times$ to 1.4$\times$ due to the overhead of repeatedly recomputing pruned tokens (i.e., skipped tokens). These results show that overly infrequent reselection harms quality, whereas overly frequent reselection yields diminishing returns but higher cost; thus, we use 5 reselections (every 10 blocks in Hunyuan-Video model) as our default setting.

\begin{table}[h]
\centering
\small
\setlength{\tabcolsep}{3pt}
\caption{Sensitivity of TAPE to the smoothing weight $\alpha$. In this experiment, we set the token reduction rate to 30\%.
We compare \textit{Speedup} (higher is better), total VBench score, and three key dimensions where attention-only baselines degrade most. 
}
\label{tab:eval_sens_alpha}
\resizebox{\textwidth}{!}{
\begin{tabular}{m{2.8cm}<{\centering} m{1.6cm}<{\centering} m{1.8cm}<{\centering} m{3.6cm}<{\centering} m{3cm}<{\centering} m{2.6cm}<{\centering}}
\toprule
\multirow{2}{*}{Method} & \multirow{2}{*}{Speedup $\uparrow$} & \multirow{2}{*}{Total Score $\uparrow$} & \multicolumn{3}{c}{VBench Dimensions $\uparrow$} \\
\cmidrule(lr){4-6}
& & & Background Consistency & Temporal Flickering & Imaging Quality \\
\midrule
Baseline (no pruning)                           & 1.0x & 83.24 & 97.76 & 99.44 & 67.56 \\
\midrule
TAPE ($\alpha{=}0.1$)                           & 1.5x & 80.67 & 96.33 & 98.68 & 64.52 \\
TAPE ($\alpha{=}0.3$)                           & 1.5x & 81.64 & 96.35 & 98.43 & 65.48 \\
\rowcolor[gray]{.9} TAPE ($\alpha{=}0.5$)       & 1.5x & 82.01 & 96.23 & 98.36 & 66.03 \\
TAPE ($\alpha{=}0.7$)                           & 1.5x & 81.45 & 95.67 & 97.90 & 66.11 \\
TAPE ($\alpha{=}0.9$)                           & 1.5x & 81.22 & 95.32 & 97.73 & 66.16 \\
\bottomrule
\end{tabular}
}
\end{table}

\begin{table}[h]
    \setlength\tabcolsep{2pt}
	\centering
	\caption{Sensitivity study on token reselection frequency/times. In this experiment, the Hunyuan-Video model (with 60 attention blocks in total) is used as the base model. The token reduction rate is set to 30\%.}
    \label{tab:eval_sens_freq}
    \resizebox{\textwidth}{!}{
	\begin{tabular}{m{4.2cm}<{\centering} m{2.2cm}<{\centering} m{2.4cm}<{\centering} m{2.6cm}<{\centering} m{2.6cm}<{\centering}}
	\toprule
	    Reselection Times (Frequency)                  & Speedup $\uparrow$   & Total Score $\uparrow$      & Quality Score $\uparrow$  & Semantic Score $\uparrow$ \\
		\midrule
		  10 Times (every 5 blocks)                           & 1.4x                     & 82.04                 & 83.68          & 75.48 \\
        \rowcolor[gray]{.9} 5 Times (every 10 blocks)       & 1.5x                     & 82.01                  & 83.64                      & 75.50  \\
        3 Times (every 15 blocks)                           & 1.5x                     & 81.51                  & 83.26                      & 74.53  \\
        2 Times (every 20 blocks)                           & 1.5x                     & 81.27            & 83.07   & 74.08 \\
        \bottomrule
	\end{tabular}
    }
\end{table}

\section{Evaluation when using Wan as Base Model}
\label{supp:result_wan}
In this section, we use Wan2.1-T2V-1.3B~\citep{wan2025wan} as the base model to evaluate the performance of our proposed TAPE pruning framework.
As shown in Table~\ref{tab:wan_results}, under the same token reduction rate, TAPE achieves comparable speedup to other pruning methods but consistently obtains the highest Total, Quality, and Semantic Scores among all pruning methods, indicating the highest video quality. 
Notably, at the 20\% token reduction rate, TAPE’s VBench scores are almost unchanged compared to the baseline without pruning, indicating that it largely preserves video quality while still providing acceleration. 
These results demonstrate the effectiveness of TAPE in delivering efficient token pruning with minimal quality loss, even when applied to a different base model.

\begin{table}[!t]
    \setlength\tabcolsep{3pt}
	\centering
	\caption{Results on VBench. In this experiment, Wan2.1-T2V-1.3B is used as the base model.}
    \resizebox{\textwidth}{!}{
	\begin{tabular}{m{2.4cm}<{\centering} m{2.2cm}<{\centering} m{2.2cm}<{\centering} m{2.6cm}<{\centering} m{2.6cm}<{\centering} m{2.6cm}<{\centering}}
	           \toprule
               \multirow{2}{*}{\makecell{Token \\ Reduction Rate}}  & Method & Speedup $\uparrow$ & Total Score $\uparrow$ & Quality Score $\uparrow$ & Semantic Score $\uparrow$  \\ 
                        \cline{2-6}
                                                & Baseline                            & 1.0x  & 83.31 & 85.23 & 75.65  \\ 
                                                \hline
        \multirow{4}{*}{ 20\%}                  & EViT           & 1.3x  & 81.88 & 84.04 & 73.22  \\ 
                                                & ToMe         & 1.3x  & 80.93 & 83.22 & 71.76  \\ 
                                                & STA-ViT        & 1.2x  & 81.60 & 83.61 & 73.54  \\ 
                        \rowcolor[gray]{.9}  &  TAPE (Ours)                           & 1.3x  & 83.27 & 85.20 & 75.53  \\ 
                                                \hline
        \multirow{4}{*}{30\%}                     & EViT         & 1.5x  & 80.34 & 82.28 & 72.60  \\ 
                                                    & ToMe     & 1.5x  & 79.30 & 81.54 & 70.34  \\ 
                                                    & STA-ViT    & 1.4x  & 79.16 & 81.05 & 71.62  \\ 
                                            \rowcolor[gray]{.9}    &  TAPE (Ours)     & 1.5x  & 82.52 & 84.48 & 74.67  \\ 
                                                \hline
        \multirow{4}{*}{40\%}                  & EViT            & 1.8x  & 78.44 & 80.35 & 70.78  \\ 
                                            & ToMe             & 1.7x  & 77.93 & 80.03 & 69.54  \\ 
                                              & STA-ViT          & 1.7x  & 77.90 & 79.63 & 70.98  \\
                                            \rowcolor[gray]{.9}   &  TAPE (Ours)     & 1.8x  & 81.58 & 83.76 & 72.85  \\ 
        \bottomrule
    \end{tabular}
    }
\label{tab:wan_results}
\end{table}

\begin{table}[htbp]
	\centering
	\caption{Comparison of different token reduction methods using the FVD metric. In this experiment, Hunyuan-Video is used as the base model.}
    \label{tab:more_metrics}
    {
	\begin{tabular}{m{3cm}<{\centering} m{2.6cm}<{\centering} m{2cm}<{\centering} m{2.2cm}<{\centering}}
	    \toprule
        Token Reduction Rate  & Method & Speedup $\uparrow$ & FVD $\downarrow$ \\
        \cline{1-4}
        \multirow{4}{*}{20\%}                  
            & EViT           & 1.3x  & 496.2 \\ 
            & ToMe         & 1.2x  & 522.9 \\ 
            & STA-ViT        & 1.2x  & 559.5 \\ 
        \rowcolor[gray]{.9}    & TAPE (Ours)     & 1.3x  & 468.7 \\ 
        \hline
        \multirow{4}{*}{30\%}                     
            & EViT           & 1.5x  & 543.6 \\ 
            & ToMe         & 1.4x  & 559.1  \\ 
            & STA-ViT        & 1.4x  & 586.8 \\ 
         \rowcolor[gray]{.9}    & TAPE (Ours)     & 1.5x  & 489.3 \\ 
        \hline
        \multirow{4}{*}{40\%}                     
            & EViT           & 1.8x  & 578.4 \\ 
            & ToMe         & 1.8x  & 597.2  \\ 
            & STA-ViT        & 1.7x  & 628.9 \\ 
         \rowcolor[gray]{.9} & TAPE (Ours)     & 1.8x  & 516.1 \\ 
        \bottomrule
    \end{tabular}
    }
\end{table}

\section{Evaluation on FVD Metric}
\label{supp:results_metrics}
In this section, we evaluate our pruning framework TAPE using the metric of Fréchet Video Distance (FVD) ~\citep{unterthiner2018towards}.
We use the text prompts from VBench to generate videos.
For each prompt, we first generate a reference video with the Hunyuan-Video model without pruning.
We then generate videos with all pruning methods using the same prompts and five random seeds (reuse the same set of seeds across all methods).
We treat the videos generated by the Hunyuan-Video without pruning as the reference, and calculate FVD for each pruning method with respect to this unpruned baseline.
As shown in Table~\ref{tab:more_metrics}, TAPE consistently outperforms the other three token reduction methods by yielding the lowest FVD with similar speedup. These results further confirm that TAPE better preserves the visual fidelity of the base model compared to other token reduction approaches.

\section{Visualization on Pruned Areas}
\label{supp:visual_prune}
In this section, we use a representative example to visualize the pruned areas (20\% token reduction) across multiple frames for both our TAPE method and EViT, where EViT prunes tokens purely based on attention scores. These results are obtained from the first transformer layer at the final denoising step. As shown in Figure \ref{fig:visual_pruned_areas}, both TAPE and EViT mainly remove less-informative background regions. However, our temporal-aware pruning produces much more coherent masks across frames. For example, if we take a closer look at the top-right pruned regions inside the red box, EViT’s masks fluctuate noticeably from frame to frame, whereas TAPE maintains a stable and coherent pruning pattern. This helps reduce frame-wise flickering.

\begin{figure}[h]
    \centering
    \includegraphics[width=0.9\textwidth]{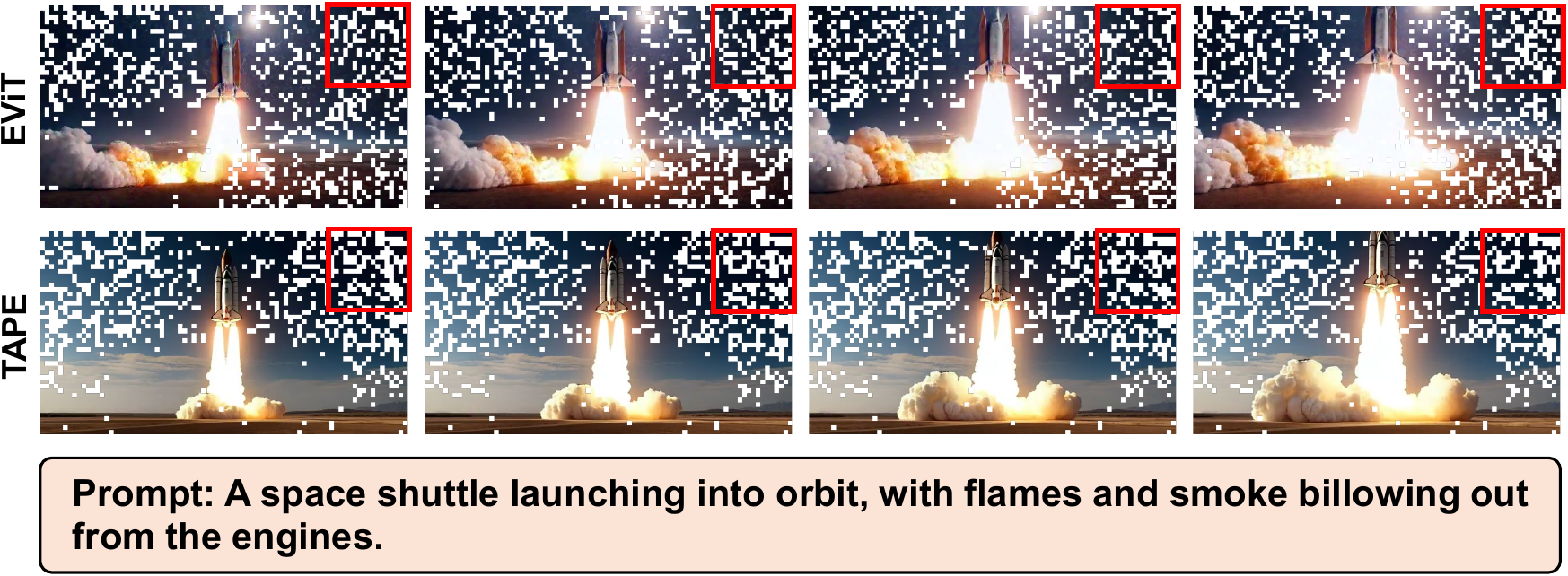}
    \caption{An example visualization of pruned areas across frames for EViT and our proposed TAPE with 20\% token reduction.}
    \label{fig:visual_pruned_areas}
\end{figure}

\section{Detailed Results of all VBench Dimensions}
\label{supp:result_allD}
In this section, we present the detailed VBench results for all evaluation dimensions corresponding to Table \ref{tab:main_results} in the main paper. Specifically, the scores of all quality-related dimensions are reported in Table \ref{tab:vbench_quality_dims}, while the scores of all semantic-related dimensions are reported in Table \ref{tab:vbench_semantic_dims}.

\begin{table}[!htbp]
    \scriptsize
    \setlength\tabcolsep{3pt}
    \centering
    \caption{VBench quality scores in all evaluation dimensions under different token reduction methods.}
    \label{tab:vbench_quality_dims}
    \resizebox{\textwidth}{!}{
    \begin{tabular}{m{2cm}<{\centering} m{2cm}<{\centering}
                    m{1.6cm}<{\centering} m{1.7cm}<{\centering}
                    m{1.7cm}<{\centering} m{1.7cm}<{\centering}
                    m{1.5cm}<{\centering} m{1.5cm}<{\centering} m{1.5cm}<{\centering}}
        \toprule
        \multirow{3}{*}{\makecell{Token \\ Reduction Rate}} & \multirow{2}{*}{Method} & \multirow{2}{*}{\makecell{Subject \\ Consistency}} & \multirow{2}{*}{\makecell{Background \\ Consistency}} & \multirow{2}{*}{\makecell{Temporal \\ Flickering}} & \multirow{2}{*}{\makecell{Motion \\ Smoothness}} & \multirow{2}{*}{\makecell{Aesthetic \\ Quality}} & \multirow{2}{*}{\makecell{Imaging \\ Quality}} & \multirow{2}{*}{\makecell{Dynamic \\ Degree}} \\ \\
        \cline{2-9}
        & Baseline & 97.37 & 97.76 & 99.44 & 98.99 & 60.36 & 67.56 & 70.83 \\
        \midrule
        \multirow{4}{*}{20\%}
            & EViT     & 96.89 & 96.08 & 98.22 & 98.78 & 59.83 & 65.64 & 70.28 \\
            & ToMe     & 97.45 & 96.06 & 98.20 & 98.61 & 59.98 & 66.60 & 71.94 \\
            & STA-ViT  & 96.77 & 94.78 & 97.30 & 98.56 & 59.52 & 65.20 & 70.28 \\
          \rowcolor[gray]{.9}  & TAPE (Ours)      & 97.36 & 97.55 & 99.33 & 98.95 & 60.13 & 67.36 & 70.56 \\
        \midrule
        \multirow{4}{*}{30\%}
            & EViT     & 96.40 & 95.24 & 97.66 & 98.83 & 59.23 & 64.98 & 70.28 \\
            & ToMe     & 96.53 & 94.95 & 97.42 & 98.76 & 58.94 & 65.27 & 69.72 \\
            & STA-ViT  & 96.45 & 92.80 & 96.30 & 98.41 & 58.68 & 63.64 & 69.17 \\
          \rowcolor[gray]{.9}  &  TAPE (Ours)      & 96.47 & 96.23 & 98.36 & 98.90 & 59.10 & 66.03 & 70.28 \\
        \midrule
        \multirow{4}{*}{40\%}
            & EViT     & 95.72 & 93.48 & 96.82 & 98.53 & 58.28 & 63.38 & 69.72 \\
            & ToMe     & 95.02 & 91.07 & 95.63 & 98.30 & 57.30 & 62.16 & 68.89 \\
            & STA-ViT  & 95.35 & 90.77 & 94.87 & 98.22 & 57.63 & 61.54 & 68.33 \\
         \rowcolor[gray]{.9}   &  TAPE (Ours)      & 95.88 & 95.75 & 98.11 & 98.82 & 58.77 & 65.79 & 70.28 \\
        \bottomrule
    \end{tabular}}
\end{table}

\begin{table}[H]
    \scriptsize
    \setlength\tabcolsep{3pt}
    \centering
    \caption{VBench semantic scores in all evaluation dimensions under different token reduction methods.}
    \label{tab:vbench_semantic_dims}
    \resizebox{\textwidth}{!}{
    \begin{tabular}{m{2cm}<{\centering} m{2cm}<{\centering}
                    m{1cm}<{\centering} m{1cm}<{\centering}
                    m{1cm}<{\centering} m{0.9cm}<{\centering}
                    m{1.5cm}<{\centering} m{1cm}<{\centering}
                    m{1.4cm}<{\centering} m{1.4cm}<{\centering}
                    m{1.4cm}<{\centering}}
        \toprule
        \multirow{3}{*}{\makecell{Token \\ Reduction Rate}} & \multirow{2}{*}{Method} 
        & \multirow{2}{*}{\makecell{Object \\ Class}} 
        & \multirow{2}{*}{\makecell{Multiple \\ Objects}} 
        & \multirow{2}{*}{\makecell{Human \\ Action}} 
        & \multirow{2}{*}{Color} 
        & \multirow{2}{*}{\makecell{Spatial \\ Relationship}} 
        & \multirow{2}{*}{Scene} 
        & \multirow{2}{*}{\makecell{Appearance \\ Style}} 
        & \multirow{2}{*}{\makecell{Temporal \\ Style}} 
        & \multirow{2}{*}{\makecell{Overall \\ Consistency}} \\ \\
        \cline{2-11}
        & Baseline & 86.10 & 68.55 & 94.40 & 91.60 & 68.68 & 53.88 & 19.80 & 23.89 & 26.44 \\
        \midrule
        \multirow{4}{*}{20\%}
            & EViT        & 85.02 & 67.20 & 93.20 & 90.31 & 66.78 & 53.52 & 19.60 & 23.62 & 25.75 \\
            & ToMe      & 85.34 & 67.80 & 94.20 & 91.10 & 68.65 & 53.38 & 19.54 & 23.76 & 25.98 \\
            & STA-ViT     & 83.51 & 66.23 & 91.42 & 88.07 & 66.90 & 52.40 & 19.20 & 22.83 & 25.45 \\
          \rowcolor[gray]{.9}  & TAPE (Ours)  & 85.90 & 68.44 & 95.20 & 92.55 & 69.87 & 54.08 & 19.87 & 24.03 & 26.43 \\
        \midrule
        \multirow{4}{*}{30\%}
            & EViT        & 83.31 & 66.75 & 91.60 & 89.53 & 66.82 & 52.94 & 19.39 & 23.15 & 25.96 \\
            & ToMe      & 84.23 & 66.51 & 91.40 & 89.38 & 66.24 & 52.08 & 19.04 & 23.41 & 25.82 \\
            & STA-ViT     & 81.28 & 64.90 & 88.80 & 86.23 & 64.00 & 50.14 & 18.70 & 22.42 & 24.90 \\
          \rowcolor[gray]{.9}  & TAPE (Ours)  & 86.27 & 68.58 & 94.20 & 91.41 & 68.18 & 53.12 & 19.72 & 23.66 & 26.31 \\
        \midrule
        \multirow{4}{*}{40\%}
            & EViT        & 80.95 & 63.58 & 87.40 & 85.17 & 63.79 & 50.52 & 18.52 & 22.47 & 24.78 \\
            & ToMe      & 81.57 & 65.60 & 89.00 & 86.16 & 64.57 & 51.47 & 18.79 & 22.49 & 25.11 \\
            & STA-ViT     & 78.64 & 63.41 & 86.80 & 84.05 & 62.90 & 49.77 & 18.17 & 21.75 & 24.08 \\
          \rowcolor[gray]{.9}  &  TAPE (Ours)  & 82.71 & 65.86 & 90.20 & 86.85 & 66.08 & 51.61 & 18.94 & 22.69 & 25.02 \\
        \bottomrule
    \end{tabular}}
\end{table}

\section{Visualization on Generated Videos}
\label{supp:visual_video}
Due to the limited space in the main paper, we provide additional visualizations of generated videos in this section, comparing the baseline Hunyuan-Video model without pruning and our TAPE under a 20\% token reduction setting.
As shown in Figure \ref{fig:visual_video_supp}, across diverse prompts, TAPE effectively preserves the semantic content and motion patterns specified by the text, while maintaining clear structures and visually detailed frames.
Moreover, the videos produced by TAPE remain visually close to the unpruned baseline in both scene composition and temporal coherence.

\begin{figure}[!htbp]
    \centering
    \includegraphics[width=0.99\textwidth]{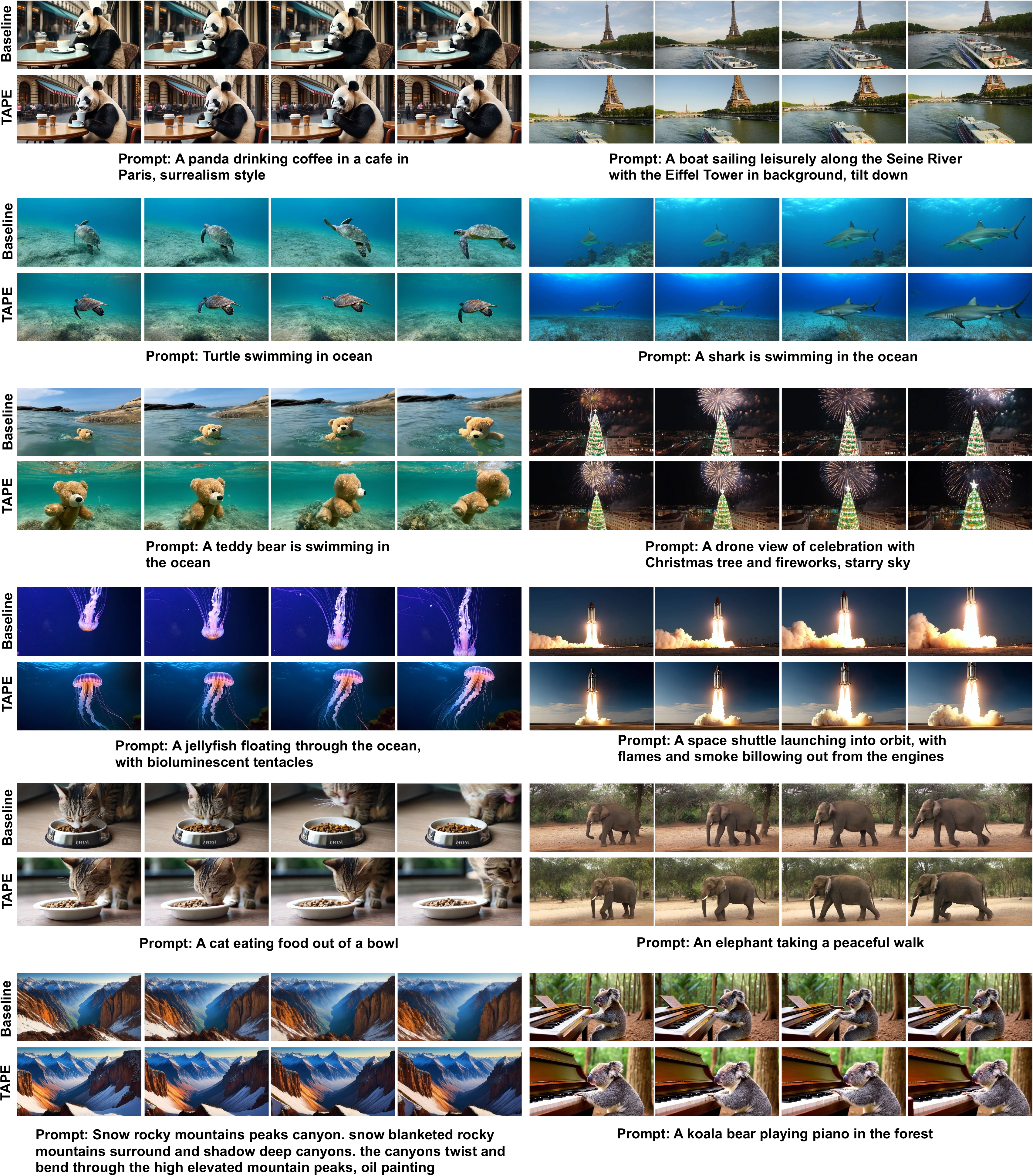}
    \caption{Additional visualizations of the generated videos.}
    \label{fig:visual_video_supp}
\end{figure}

\end{document}